\documentclass[letterpaper, 10 pt, conference]{ieeeconf}
\IEEEoverridecommandlockouts              
\usepackage{graphics}
\usepackage{mathptmx} 
\usepackage{times}
\usepackage{subfig}
\usepackage{float}
\usepackage{amssymb, amsmath, amsbsy}
\usepackage{tabularx}
\usepackage[utf8]{inputenc}
\usepackage{listings}
\usepackage{amsfonts}
\usepackage{xfrac}
\usepackage{nicefrac} 
\usepackage{booktabs}
\DeclareUnicodeCharacter{2212}{-} 

\usepackage{cite}
\usepackage{array,graphicx}
\usepackage{booktabs}
\usepackage{pifont}

\title{\LARGE \bf On the generalization capabilities of FSL methods through domain adaptation:  a case study in endoscopic kidney stone image classification}

\author{Mauricio Mendez-Ruiz$^{1}$, Francisco Lopez-Tiro$^{1}$, Jonathan El-Beze$^{2}$, Vincent Estrade$^{3}$,\\  Gilberto Ochoa-Ruiz$^{1}$, Jacques Hubert$^{2}$, Andres Mendez-Vazquez$^{4}$, Christian Daul$^{5}$\\
\thanks{$^{1}$Tecnológico de Monterrey, School of Engineering and Sciences, Mexico}%
\thanks{$^{2}$CHU Nancy, Service d'urologie de Brabois, F-54511 Nancy, France}%
\thanks{$^{3}$CHU Pellegrin place Amémlie Raba Léon F-33000 Bordeaux, France}%
\thanks{$^{4}$Centro de Investigación y de Estudios Avanzados del Instituto Politécnico Nacional, Mexico}%
\thanks{$^{5}$CRAN (Universit\'e de Lorraine and CNRS), F-54000 Nancy, France}%
\thanks{Corresponding Authors: gilberto.ochoa@tec.mx, christian.daul@univ-lorraine.fr}}%

\begin{document}
\maketitle
\thispagestyle{empty}
\pagestyle{empty}
\begin{abstract}

Deep learning has shown great promise in diverse areas of computer vision, such as image classification, object detection and semantic segmentation, among many others.
However, as it has been repeatedly demonstrated, deep learning methods trained on a dataset do not generalize well to datasets from other domains or even to similar datasets, due to data distribution shifts.
In this work, we propose the use of a meta-learning based few-shot learning approach to alleviate these problems. In order to demonstrate its efficacy, we use  two datasets of kidney stones samples acquired with different endoscopes and different acquisition conditions. The results show how such methods are indeed capable of handling domain-shifts by attaining an accuracy of 74.38\% and 88.52\% in the 5-way 5-shot and 5-way 20-shot settings respectively. Instead, in the same dataset, traditional Deep Learning (DL) methods attain only an accuracy of 45\%.
\end{abstract}

\section{Introduction}


Progresses made in Artificial Intelligence (AI) in recent years show great results that compare or surpass human capabilities in a set of problems such as natural language processing \cite{nlp2}, smart agriculture \cite{agriculture1}, object recognition \cite{objectdetection3}, healthcare and medical applications \cite{healthcare1}, among others. Such AI-based models have been possible  due to the existence of  large scale labelled datasets enabling to extract profound knowledge about the data. However, it has been demonstrated that if the same models are deployed in slightly different operating conditions (i.e., medical imaging applications) such AI methods are in fact are very fragile \cite{general2}, as they exhibit very poor generalization properties
\cite{general1}. Such variations can stem from changes in acquisition devices and/or the operating conditions in clinical settings, which can hamper the adoption of AI-based CAD tools in many applications \cite{general3}.
To make matters worse, most DL methods in the state of the art still require humongous amounts of data to be trained, which is not realistic in most of the medical application domains.
Therefore, in recent years, the meta-learning and Few-Shot Learning (FSL) paradigms have emerged as a means to cope with the training data scarcity problem and furthermore, to make models more capable of generalization with less computing effort  and incremental learning capabilities \cite{general4}.

\begin{figure*}
\begin{center}
\includegraphics[width=\textwidth]{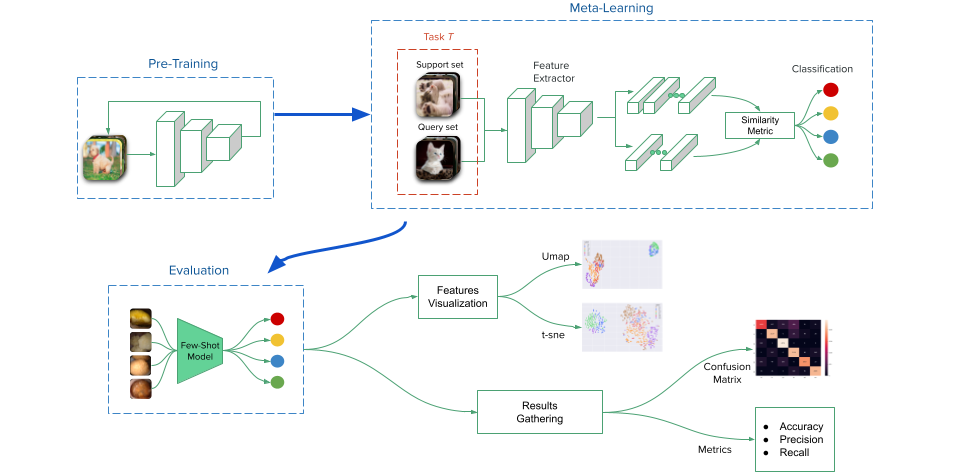}
\end{center}
   \caption{The proposed model is divided into three stages: pre-training, meta-learning and evaluation.}
\label{fig:model}
\end{figure*}

In this work, we propose a novel meta learning-based few shot learning approach for image classification to assess the generalization of a trained model in two different kidney stones datasets. Our method is based on the following two core ideas: i) The use of a pre-trained model that learns representations through self-supervised learning (instead of a supervised one) can enhance the features generalization and ii) a meta-learning stage can be used to further fine-tune the model for specific domains in order to improve its performance. The advantage of our method is that, compared to other DL approaches to classify kidney stones, it can better generalize to other data distributions and obtain good results without the need of manual data augmentation.
In order to validate our proposal, we make use of two datasets of kidney stones acquired using endoscopes of two different vendors and of different technical characteristics and under different acquisition conditions (in-vivo, ex-vivo). As illustrated in Figure \ref{fig:model} our method is divided in three stages. First, in the pre-training phase, we use a self-supervised model for the embedding network. Then, we proceed to a meta-learning stage, where we fine-tune our model. Lastly, we evaluate the model with the two kidney stones datasets and obtain the corresponding metrics and features visualization, which are of great help to better assess the generalization capabilities of the model.

The rest of the paper is organized as follows. In section 2 we present the medical motivation for this work, while section 3 discusses the state of the art of FSL. In section 4 we introduce the proposed method and in Section 5 we provide details about the implementation and experiments made in order to obtain the best model, as well as the comparison with previous models. Finally, in section 6 we summarize our work and discuss some perspectives for future work.

\section{Medical Context and Motivation}

In recent years, there has been an increased interest in the recognition of kidney stones morphologies (i.e. crystalline type) for speeding the diagnosis and treatment processes \cite{laube2020, jah2020}. The traditional (en-vivo) approach, known in the medical field as morpho-constitutional analysis (MCA \cite{estrade2020, corrales2021}) , relies on an inspection of the surface and section of the images under the microscope, followed by a SPIR analysis to determine their biochemical composition (i.e. stone type). This analysis is essential as it provides very important information about the lithogenesis (i.e. cause of formation) of the stone, but modern extraction techniques increasingly rely on a technique for pulverizing the stone. Such technique, known as dusting, leads to a destruction of the morphological information of the sample or even an alteration of its biochemical composition \cite{keller2018}, making it impossible to prescribe an appropriate and timely treatment.

Therefore, specialists have sought solutions in the form image classification methods for categorizing kidney stone samples, first using ex-vivo images corpuses \cite{serrat2017, black2020} and later tackling the problem of endoscopy stone recognition (ESR) \cite{embc2020, embc2021, estrade2021}. Although the results have been encouraging (up to 98\% average precision for pure stones), some of the methodological choices make the results far from conclusive as more complete studies are needed \cite{sommen2020}.

First, the majority of ML-based ESR methods make use of very small datasets and rely heavily on patch sampling \cite{embc2021, black2020}, which might introduce bias towards certain classes (i.e. data leakage). Furthermore, most datasets reported in the literature contain only a small fraction of the 21 identified classes of kidney stones (up to 6 classses) which might yield overly optimistic results \cite{Aggarwal2021}. Another aspect is that most of the existing methods have been tested on images acquired using one or at most two endoscopes types from the same hospital, which can lead to problems such as shortcut learning, or simply the samples might not be representative enough of the underlying distribution. This leads to a problem increasingly reported in the literature: models trained with data from certain acquisition devices or under certain imaging circumstances do not generalize well to data from a different distribution \cite{yao2019}. In fact, we show later in the article that models trained on one dataset present a significant drop in performance when tested with unseen data (i.e. the same classes acquired with a different ureterescope).

In order to address some of the issues mentioned above, in this work we explore recent developments in the DL field, namely few shot learning and meta-learning strategies that are promising areas of research for training models with few samples and capable of better generalization capabilities. For validating our approach we make use of two very distinct kidney stones datasets, containing data from the same classes, but different distributions (i.e., the data was acquired with different endoscopes and under different acquisition conditions). To the best of our knowledge, our work is of the firsts to assess the generalization capabilities of ML models applied on  endoscopic images.

\section{Related Work}

\subsection{Few-shot learning}

Research on FSL has received an increased attention on recent years, as it has continuously demonstrated good results on problems with low availability of data \cite{prototypicalNets2017, matchingNetworks2016}. Recent models for FSL adopt a meta-learning strategy, an  approach thats seeks to learn discriminant features across tasks, later adapting the model to new tasks.


We can categorize FSL approaches into two main branches: 1) metric-learning based, and 2) optimization based. The goal of metric-learning approaches is to learn a similarity metric expected to generalize across different tasks. There are baseline methods which have achieved important milestones for FSL, such as Prototypical Networks \cite{prototypicalNets2017}, Matching Networks \cite{matchingNetworks2016} and Relation Networks \cite{sung2017relationNet}. Optimization based approaches make use of a base-learner and a meta-learner, where the meta-learner's parameters are optimized by gradual learning across tasks to promote a faster learning of the base-learner for each specific task. The Model-Agnostic Meta-Learning (MAML) approach \cite{maml} was the first one to use this strategy to parameter initialization, such that the base learner can rapidly generalize from an initial guess of parameters. Thus, extensions to MAML have been proposed \cite{reptile, optimizationfewshot} to improve the optimization of the meta-learner.

\subsection{Cross-domain FSL}

Domain adaptation refers to the transfer of knowledge from one or multiple source domains to a target domain with a different data distribution. 
Several approaches have been proposed to address this issue: discrepancy-based models \cite{domain_adapt1, domain_adapt2}, adversarial-based methods \cite{domain_adapt3, domain_adapt4} and reconstruction-based approaches \cite{domain_adapt5, domain_adapt6}. Nevertheless, even when the data distributions are different, these methods operate in situations where the training and test sets contain the same classes.

For cross-domain FSL, where base and novel classes come from different domains, this can introduce non-desirable variations in the performance of the models. In \cite{closer_look_fsl}, the authors made an analysis of different meta-learning methods in the cross-domain setting. They proposed a cross-domain scenario which trains on miniImagenet dataset and test with CUB dataset images. Further on, in \cite{broader_study_cross_fsl}, the authors 
established a challenging benchmark consisting of images of diverse types with an increasing dissimilarity degree to natural images and with various levels of perspective distortion, semantic content and color depth. They also evaluated the performance of existing meta-learning methods on this benchmark. All these works show how cross-domain paradigm is able to enhance the results the base models,  making cross-domain approaches one of the corner stones of FSL

\subsection{Kidney Stones Classification}

Different approaches have been proposed to deal with the classification of kidney stones from still images, acquired through different means. For instance, authors in \cite{serrat2017} made use of ex-vivo images to train a Random Forest classifier, which exploits histograms of RGB colours and local binary patterns. As a continuation of this work, they trained a DL methods base on a Siamese CNN \cite{torrell} using the same dataset. Although such methods showed the potential of image-based recognition of kidney stones, the proposed models obtained a moderate performance of 71\% and 74\% on mean accuracy, respectively. The authors in \cite{black2020} improved these results by using a ResNet-101 and employing a data augmentation technique to leverage the small dataset. These previous methods have the limitation of being tested on ex-vivo images obtained in highly controlled acquisition conditions, while in-vivo images are more difficult to utilize due to the complex acquisition conditions present in endoscopic procedures \cite{juan_carlos} . For in-vivo images, authors in \cite{embc2020} used classical classifiers such as Random Forest and kNN ensemble models obtaining an improved accuracy of 85\% over 3 kidney stone classes. Later on, \cite{embc2021} used CNN-based models to further improve their results, obtaining an accuracy of over 90\% on precision and recall over 4 classes.

Although these kidney stone classification methods have yielded increasingly good performances, they pose certain problems. First, they all require a great deal of manual data augmentation work to create several patches from the images. Second, the training and evaluation has been made over those patches instead of the full images, which is not realistic. Third, all these previous methods have shown poor generalization capabilities, as they are trained with images from specific data acquisition conditions and fail to classify when using a different acquisition method, as we will discuss in more detail in a subsequent section.



\section{Proposed Approach}

FSL is a growing research field with the challenging problem of learning from limited data, while verifying the performance of the models on previously unseen classes. Additional to this, FSL methods are expected to generalize well to other data distributions. In this work, as shown in Figure \ref{fig:model}, we follow a two stages paradigm (pre-training and meta-learning) to train models capable of generalizing to a different data distribution to classify unseen kidney stones classes. In this approach, the model is first pre-trained on a large datasets of natural images (i.e., ImageNet) to obtain a good initial estimate of the model parameters. Then the model is fine-tuned using a meta-learning approach with increasingly similar datasets to the target domain. Finally, the model is tested with data coming from a completely new distribution to assess it genelization capabilities.


\subsection{Datasets}

\subsubsection{Base Datasets}

The MiniImagenet dataset \cite{matchingNetworks2016} is a subset of the ImageNet dataset \cite{imagenet}. It is comprised of 100 classes with 600 images per class, making up a total of 60,000 images. 

This dataset is widely used in the FSL literature for image classification \cite{maml, prototypicalNets2017}, following the same split proposed in \cite{matchingNetworks2016} by sampling images of 64 classes for training, 16 classes for validation and 20 classes for testing.

Caltech-UCSD Birds-200-2011 (CUB-200-2011) \cite{cub} is an image dataset with more specialization in the categories, containing photos of 200 bird species.This dataset was created for the study of subordinate categorization, which is not possible with other datasets like ImageNet. Most few-shot learning approaches in the literature make use of CUB-200 for measuring its performance, following the protocol proposed by \cite{fsl_cub_split}.


\subsubsection{Cross-Domain datasets}

For the cross-domain adaptation in FSL, we used the datasets suggested by \cite{cross-domain}, comprised of images from multiple domains (i.e., image modalities and acquisition conditions). The first dataset is CropDiseases \cite{cropdiseases}, which contains 38 classes with images of plant leaves. We split the dataset into 30 classes for training and 8 for validation. The second dataset is EuroSAT \cite{eurosat}, which contains satellite images for land use and land cover classification. It is comprised of 10 classes that we split into 5 classes for training and 5 for validation. The third dataset is ISIC \cite{isic}, containing skin diseases images for skin image analysis. It contains 7 categories, which we split by using 5 for training and 2 for testing. The fourth dataset is Chest-X, containing X-ray chest images for the diagnosis of many lung diseases.

\subsubsection{Kidney Stones datasets}
\label{ref: ks_datasets}

\begin{figure*} [h] 
    \centering
    \subfloat[In-vivo images \label{fig:2a}]{
    \label{fig:vincent-estrade_imgs}\includegraphics[width=0.32\linewidth]{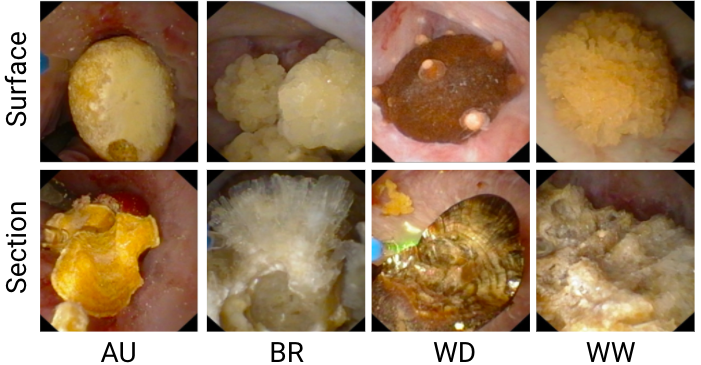}}
    \hspace{5mm}
    \subfloat[Ex-vivo images \label{fig:2b}]{
    \label{fig:jonathan-elbeze_imgs}\includegraphics[width=0.48\linewidth]{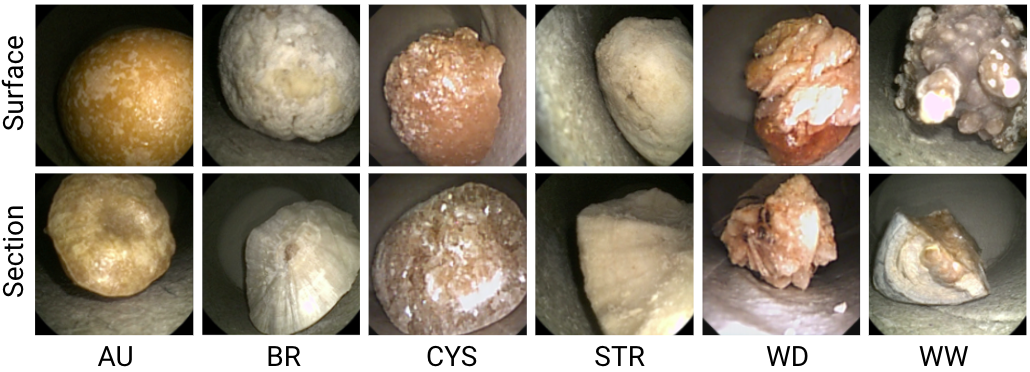}}
    \caption{Examples of kidney stone images from the (a) in-vivo and  (b) ex-vivo dataset. The categories shown are, from left to right, uric acid (AU), brushite (BR), cystine (CYS), struvite (STR), weddelite (WD) and whewellite (WW). Surface images are on the top row and section images are on the bottom row.}
    \label{fig:total2}
    \end{figure*}

We make use of two different kidney stones datasets, obtained using two different acquisition devices (i.e., endoscopes) and under divergent acquisition conditions and lighting environments: the first is comprised by in-vivo images, whilst the second was created by capturing images with an endoscope in ex-vivo conditions, see Fig. \ref{fig:total2}. 

The in-vivo dataset includes 156 kidney stone images acquired in-vivo (i.e. during actual ureteroscopic interventions) and which were annotated by expert Dr. Vincent Estrade (an urologist involved in MCA). The dataset consists of 65 cross-section images and 91 surface images from the 4 classes of kidney stones with the highest incidence: uric acid (AU), brushite (BR), weddelite (WD) and whewellite (WW). Table \ref{tab:vincent-estrade_ds} details the amount of images from each kidney stone category. The images from this dataset were captured using different ureteroscopes from Olympus and the Richard Wolf company. Some images from this dataset are shown on Fig. \ref{fig:2a}. It must be noted that such images are more challenging that those present in other datasets, which make use of high resolution images acquired ex-vivo using either professional cameras or digital microscopes. More details can be found in \cite{estrade2020}. 

The ex-vivo dataset consists of 765 kidney stones images, with 318 corresponding to cross-section and 447  corresponding to surface images, from 6 of the kidney stones categories with higher incidence: uric acid (AU), brushite (BR), cystine (CYS), struvite (STR), weddelite (WD) and whewellite (WW). The ex-vivo columns in Table \ref{tab:vincent-estrade_ds} detail the amount of images and their type in this dataset. Figure \ref{fig:jonathan-elbeze_imgs} show examples of images from this dataset. The images were acquired with an actual endoscope from Karl Storz, and great care has been taken to reproduce the conditions of the ureteroscopic interventions in terms of the tissue sorrounding the stones and the field of view, as well as possible photometric artifacts that can affect the image quality. More detials about the acsquision process and characteristics of this dataset can be found in \cite{elbeze2022}

\begin{table}[]
\scalebox{1}{\begin{tabular}{@{}rcccccc@{}}
\cmidrule(l){2-7}
 & \multicolumn{3}{c}{\textbf{In-vivo}} & \multicolumn{3}{c}{\textbf{Ex-vivo}} \\ \midrule
\textbf{Class/View} & \textbf{Section} & \textbf{Surface} & \textbf{Total} & \textbf{Section} & \textbf{Surface} & \textbf{Total} \\ \midrule
Acid Uric & 17 & 17 & 34 & 62 & 70 & 132 \\
Brushite & 12 & 12 & 24 & 95 & 109 & 204 \\
Cystine & --- & --- & --- & 53 & 52 & 105 \\
Struvite & --- & --- & --- & 28 & 52 & 80 \\
Weddelite & 9 & 32 & 41 & 48 & 85 & 133 \\
Whewellite & 27 & 30 & 57 & 32 & 79 & 111 \\ \midrule
Total & 65 & 91 & 156 & 318 & 447 & 765 \\ \bottomrule
\end{tabular}}
\caption{Number of acquired images from the in-vivo and ex-vivo kidney stones dataset.}
\label{tab:vincent-estrade_ds}
\end{table}

\subsection{Model training and design choices }

\subsubsection{Pre-training stage}

The first approaches for FSL used a simple ConvNet backbone, made up of 4 convolutional blocks, as the feature extractor \cite{prototypicalNets2017}, leading to good results that supported the promising research of these methods. Current state-of-the-art models with high performance make use of a ResNet-12 \cite{categoryTraversal} or a wide ResNet \cite{predicting_params}, outperforming other models which make use of deeper embedding networks. This could be because, since we access to a very small amount of samples, deeper networks are more prone to overfitting in the supervised training. Therefore, an essential element for the meta-learning process is the initial embedding network. Several state-of-the-art methods \cite{ssl, meta-baseline} demonstrate that having a pre-trained network allows us to use a deeper backbone, thus greatly improving the performance of few-shot learning models.

Among the pre-training strategies for FSL available in the literature, a self-supervised learning  (SSL) method \cite{ssl} stands out, as it has demonstrated a good results in this taks, outperforming other state-of-the-art models. We adopt this pre-training strategy, which allows us to use a large embedding network. The SSL stage works as follows: using the Augmented Multiscale Deep InfoMax (AMDIM) model, it optimizes the network by looking for mutual local and global features from different views of an instance, thus enhancing the feature's generalization for new tasks. This makes the trained network more transferable to different domains and data distributions.

\subsubsection{Meta-learning stage}
\label{ref:meta-learning-stage}

After finding an appropriate initialisation via SSL, we proceed with the meta-learning stage to learn to generalize across tasks. This stage is divided into two phases: Meta-training and meta-testing. A few-shot $K$-way $C$-shot image classification task is given $K$ classes and $C$ images per class. The task-specific dataset can be formulated as $D = \{D_{train}, D_{test}\}$, where $D_{train}=\{(X_i, y_i)\}_{i=1}^{N_{train}}$ denotes the classes reserved for the training phase and $D_{test}=\{(X_i, y_i)\}_{i=1}^{N_{test}}$ denotes the classes reserved for the testing phase.
For each meta-train task $T$, $K$ class labels are randomly chosen from $D_{train}$ to form a support set and a query set. The support set, denoted by $S$, contains $K \times C$ samples ($K$-way $C$-shot) and the query set, denoted by $Q$, contains $n$ number of randomly chosen samples from the $K$ classes.
The training phase uses an episodic mechanism, where each episode $E$ is loaded with a new random task taken from the training data. For the meta-test, the model is tested with a new task $T$ constructed with classes that weren't seen during the meta-train phase.

We follow a metric learning approach for the image classification component of our approach. Specifically, we implemented a prototypical networks \cite{prototypicalNets2017} model by computing the prototypes $c_k$ as the mean of embedded support samples for each class. For a class $k$, the prototype is represented by the centroid of the support embedding features, obtained as:

\begin{equation}
    c_k = \frac{1}{\left | S_k \right |} \sum_{(x_i, y_i) \in S} f(x_i),
\end{equation}

The classification is performed by finding the nearest prototype for a given embedded query point. The Euclidean distance is chosen as the distance function to find the nearest class prototype.In the proposed method, we first apply SSL to pre-train the large embedding network, followed by a number of meta-learning iterations to fine-tune the model. The iterations are made by training and validating the model with different datasets, as a way to alleviate the domain shift preventing a successful classification of kidney stones.

The datasets used for meta-training were selected because they meet the requirement of decreasing its similarity with ImageNet based on three orthogonal criteria: i) Existence of perspective distortion, ii) the semantic content and iii) the color depth. The ImageNet dataset contains natural and colored images with perspective. The CropDisease dataset contains natural and colored images with perspective. Meanwhile, the EuroSAT dataset contains natural and colored images but without perspective. In contrast, the ISIC and ChestX dataset contain medical and images with no perspective (in color and in B/W, respectively). We can place the kidney stones datasets used for this study within the spectrum of dissimilarity as ISIC, as they contain both medical and colored images with no perspective.

Therefore, the model gradually converges to a to a point in which more discriminating features are learned from the fine-tuning process.   This domain-specific fine-tuning done as follows: First, we meta-train the model with the base datasets (MiniImageNet and CUB-200); then we specialize the model domain by meta-training with the CropDisease, EuroSAT and ISIC datasets sequentially. The Chest-X dataset is not considered for the sequential meta-train due to its dissimilarity with the kidney stones datasets, which would only decrease the performance obtained.

After training and validating the model, we perform the meta-testing with the two kidney stones separately. These datasets are not used in the meta-training phase, so we can evaluate the generalization capabilities of the proposed method.

\section{Experimental Results}

\subsection{Implementation details}


For the training phase of our FSL model, we follow the same setting as other few-shot learning models \cite{ prototypicalNets2017} by learning across the 5-way 1-shot and 5-way 5-shot task settings and using 15 query samples for each class in the task. For the meta-training phase, we randomly sample 100 tasks over 200 epochs. We validate each epoch with 500 randomly constructed tasks using the classes reserved for validation. For the testing phase, we randomly construct 200 tasks which is enough for the little amount of data from the kidney stones datasets. 

For the feature extraction backbone, we tested two different embedding networks, a ConvNet and an  AmdimNet. Following the same approach as \cite{prototypicalNets2017, matchingNetworks2016}, the ConvNet is built using four layers of convolutional blocks. Each block is made up of a 3x3 convolution with 64 filters, followed by a batch normalization and a ReLU layer. The input images are resized to $84 \times 84$ and normalized.  For this network, we do not apply a pre-training process due to the small scale of the model. In the meta-learning phase, we use the Adam Optimizer with an initial learning rate of $1\times 10^{-3}$ and a step size of 20. For the AmdimNet, we follow the same setup provided by \cite{ssl} consisting of the pre-training stage and meta-learning stage. In the pre-train stage we use the Adam Optimizer with a learning rate of 0.0002 and an embedding dimension of 1536. We resize the unlabelled input images to a size of $128 \times 128$ pixels. In the meta-learning stage, the input images are resized to $84 \times 84$ pixels and we used the SGD as optimizer with an initial learning rate of $2\times 10^{-4}$, a step size of 20 and weight decay of 0.0005.

\subsection{Training details}


\newcommand*\rot{\rotatebox{90}}
\newcommand*\OK{\ding{51}}


\begin{table*} \centering
    \begin{tabular}{@{} cl*{10}l @{}}
        & & \rot{Mini80} & \rot{CUB150} & \rot{Crop-Desease} & \rot{Eurosat} 
        & \rot{ISIC} \\  \midrule 
& Pretrain     & \multicolumn{5}{c}{Meta-training}  & Details \\ \midrule
        \cmidrule{1-8}
1 & ImageNet 900 & \OK &     & \OK & \OK & \OK & All datasets (except CUB) sequentially meta-trained \\
2 & ImageNet 900 & \OK & \OK & \OK & \OK & \OK & All datasets sequentially meta-trained\\
3 & ImageNet 900 & \OK & \OK & \OK & \OK & \OK & All datasets meta-trained at once \\
4 & ImageNet 1k  &     & \OK & \OK & \OK & \OK & All datasets (except CUB) sequentially meta-trained\\
5 & ImageNet 1k  &     & \OK &     & \OK & \OK & All datasets sequentially meta-trained\\
6 & ImageNet 1k  & \OK & \OK & \OK & \OK & \OK & All datasets meta-trained at once\\
        \cmidrule[1pt]{1-8}
    \end{tabular}
\caption{Experimental setup with incrementally more similar datasets to the target domain.}
\label{tab:Experiment_blocks}
\end{table*}


Our experiments were carried out in three main steps. In the first we carried out quick tests to determine if the kidney stones classification was indeed a problem that few-shot learning could solve. We trained the base prototypical networks \cite{prototypicalNets2017} on the 5-way 5-shot and 20-way 5-shot settings, and tested the models generalization on both kidney stones datasets. The results obtained in this step were promising, with models that generalized much better than previous deep learning methods. The second step was a round of experiments to find the best hyper-parameters for the main model (AmdimNet). We tested a set of combinations for the optimizer, learning rate and step size, finding that the best combination was SGD optimizer with a learning rate of 0.0002 and step size of 20. Lastly, the third step was to carry out experiments to find the best performing model for the kidney stones classification. We explain the details below.

For the \textbf{pre-training} stage we trained two different embedding models: i) ImageNet900 (Img900) was trained using the SSL strategy from 900 classes of the ImageNet dataset (removing from ImageNet the 100 classes used in MiniImageNet) and ii) ImageNet1k (Img1k) is trained using self-supervised learning from the whole 1,000 classes of the ImageNet dataset without any label.

For the \textbf{meta-training} stage we adopted a domain-specific fine tuning to reduce the domain shifts between the source datasets and the target of kidney stones images. Since our testing phase is carried out with the kidney stones datasets, the base datasets were modified to only have training and validation classes in the following way. In MiniImageNet, we assign for training the classes previously used for training and validation, and for validation we use the classes previously used for testing, leading to a total of 80 classes for training and 20 for validation (Mini80). The same split is applied to the CUB dataset, in which we use 150 classes for training and 50 classes for validation (CUB150). We trained over 20 different models, splitting the experiments into 6 data configurations based on how the model would be trained. The meta-learning setting was either by incremental learning through different datasets or by meta-learning among all datasets at once. 

Table \ref{tab:Experiment_blocks} summarizes the experimental setup, comprised by the configurations just described above. The first three rows are models pre-trained with ImageNet900 and the last three configurations are models pre-trained with ImageNet 1k. For model 1, we sequentially meta-trained across Mini80, followed by CropDiseases, then EuroSAT and lastly ISIC. The same applies for model 2, but adding CUB150 after meta-training with Mini80. For model 3, we tested the following 4 settings with the idea of learning from different datasets at once: i) Training the model first with Mini80, followed by a training with the rest of datasets, ii) training first with Mini80, followed by CUB150, and then the rest of datasets, iii) train with all datasets, except CUB150, at once, and iv) training with all the datasets at once.  For the last 3 models in the table, we repeated the same experiments but with the model pre-trained on ImageNet 1k and removing Mini80 from the datasets used in meta-training, since those would represent classes already seen in the pre-training stage.

\subsection{Evaluation results}
In order to evaluate the generalization capabilities of the models trained using the various configurations in Table \ref{tab:Experiment_blocks}, the architectures were tested on the two kidney stones datasets described in Section \ref{ref: ks_datasets}.  

Following the cross-domain setup described in \cite{cross-domain}, we conducted experiments by testing the proposed model across three main few-shot tasks (i.e., 5-way 5-shot, 5-way 20-shot and 5-way 50-shot). For the in-vivo dataset, we do not have enough images per class to test the 5-way 50-shot setting. Thus, the experiments with this setting are excluded for the in-vivo classification experiments.

\subsubsection{Ablation Study Results}

In order to assess how the various components of our proposed approach affect the generalization results of the generated model, we carried our several ablation studies, shown in Table \ref{tab:best_results}. First, we validate the effectiveness of a traditional FSL approach to classify out of distribution datasets. Specifically, we trained the baseline prototypical networks \cite{prototypicalNets2017} on the 5-way 5-shot and 20-way 5-shot settings. After testing both models, we obtained an accuracy performance of over 70\% in ex-vivo and in-vivo images, and the models demonstrated an acceptable generalization as they do not exhibit an important loss of performance (around 2\% - 3\%) when compared on the two kidney stones datasets.

Afterwards, we tested the performance yield by the models when only pre-training the models, to assess whether or not there is a gain in performance when applying the proposed meta-training strategy. After evaluating the effectiveness of two models (Img900 and Img1k), we found that the accuracy performance greatly increased (around 12\% for the in-vivo dataset and around 7\% for the ex-vivo dataset, using the Img900 and Img1k models, respectively). Nonetheless, a decrease in the generalization capabilities of the models can be observed, as there is a difference of around 8\% in the results for both datasets.

Finally, we tested the generalization performance after implementing the meta-learning strategies described in Section \ref{ref:meta-learning-stage} . We found out that there is indeed an improvement (an average of 4\% for both datasets) over the model that was only pre-trained. The improvement on accuracy is not as large as for the ProtoNet and the  pre-trained model. The difference in accuracy between the in-vivo and ex-vivo datasets is around 8\%, which is the same difference using only pre-trained models. This means that our meta-learning approach was able to account for the domain-shift of the model, while maintaining its generalization capabilities.

The best meta-learning-based models are shown in the last row of Table \ref{tab:best_results}. We can see that these models achieved a great performance improvement over the  basic FSL models (values of 88.13\% in the in-vivo dataset and 80.54\% in the ex-vivo dataset). As we will see in the next subsection, these results are comparable with some of the shallow models in the literature \cite{embc2020, embc2021} in terms of performance, and still far below the results obtained by most deep learning models. Nonetheless. the results in generalization are far more superior. Also, while most models in the state of the and have been trained with thousands of image patches, the results discussed here make use of a few hundred images.

\begin{table}[]
\centering
\scalebox{1}{\begin{tabular}{rccccc} \cline{2-6}
 & \multicolumn{2}{c}{\textbf{In-vivo}} & \multicolumn{3}{c}{\textbf{Ex-vivo}} \\ \cline{1-6}
\multicolumn{1}{r}{\textbf{Model}} & 5-shot & \multicolumn{1}{c}{20-shot} & 5-shot & 20-shot & 50-shot \\ \hline
\multicolumn{1}{r}{ProtoNet 5w5s} & 63.31 & \multicolumn{1}{c}{71.88} & 58.75 & 65.36 & 67.85 \\
\multicolumn{1}{r}{ProtoNet 20w5s} & 63.28 & \multicolumn{1}{c}{72.72} & 60.47 & 67.85 & 70.27 \\ \hline
\multicolumn{1}{r}{Img900} & 69.68 & \multicolumn{1}{c}{84.49} & 67.05 & 74.18 & 76.46 \\
\multicolumn{1}{r}{Img1k} & 70.26 & \multicolumn{1}{c}{84.25} & 66.75 & 74.36 & 77.01 \\ \hline
\multicolumn{1}{r}{Img1k-CUB-All} & 68.38 & \multicolumn{1}{c}{85.03} & 68.46 & 77.65 & 80.09 \\
\multicolumn{1}{r}{Img900 +$^{*}$} & 70.61 & \multicolumn{1}{c}{84.53} & 69.45 & 77.72 & 79.87 \\
\multicolumn{1}{r}{Img900 + $^{**}$} & 71.13 & \multicolumn{1}{c}{85.76} & 69.25 & 77.84 & 80.27 \\
\multicolumn{1}{r}{Img900 + $^{***}$} & \textbf{74.38} & \multicolumn{1}{c}{\textbf{88.13}} & \textbf{69.56} & \textbf{78.20} & \textbf{80.54} \\ \hline
\end{tabular}}
\caption{Accuracy performance of the ablation studies on our method. $^{*}$= + Mini80 + Crop Diseases, $^{**}$ = + Mini80 + CUB + All, $^{***}$ = + All (excluding CUB).}
\label{tab:best_results}
\end{table}

\begin{figure*} [t] 
    \centering
    \subfloat[In-vivo dataset]{
    \label{fig:yonathan_imgs}\includegraphics[width=0.45\textwidth]{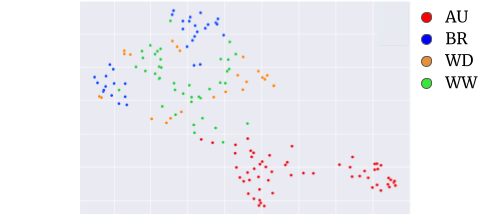}}
    \hspace{5mm}
    \subfloat[Ex-vivo dataset]{
    \label{fig:vincent-estrade_imgs}\includegraphics[width=0.45\textwidth]{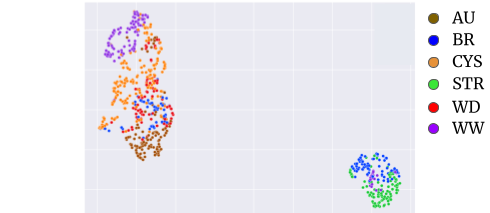}}
    \caption{UMAP feature visualization of the (a) in-vivo dataset and (b) ex-vivo dataset. The visualizations can be used in tandem with the morpho-constitutional analysis proposed by Daudon  \cite{corrales2021} to understand the errors done by the models }
    \end{figure*}

\subsubsection{Comparison with previous works}

To verify the effectiveness of our proposed method, we compare it with previous works in endoscopic stone recognition using the same datasets presented in Section \ref{ref: ks_datasets}. In  \cite{embc2021} several shallow ML methods and DL models were used to classify the images of the in-vivo dataset. We implemented some of these models and tested their generalization capabilities by trying to classify the same 4 classes of the in-vivo dataset but using the model trained on ex-vivo images. Even though the classes are the same, we obtained very poor results, showing that there is a large loss in performance (a decay of around 40\% in accuracy) when the images come from another data distribution. This issue plagues all the methods equally, but it comes to show the fragility of the existing traditional DL models.

Most of these problems can be partially fixed by using metric learning apporaches. As it can be seen in Table \ref{tab:comparison}, even the basic ProtoNet model \cite{prototypicalNets2017} is more apt at generalizing with the same performance than the classical methods mentioned before. 

Our method obtained a high performance in three metrics: precision, recall and F1-score. Compared with some of the classical machine learning methods (i.e., Random Forest, XGBoost), we obtained competitive results for the in and ex-vivo datasets. Moreover, we demonstrated that our method has much better generalization capabilities as our model exhbits a loss of 7\% on accuracy, compared with the loss of over 40\% from previous works.

\begin{table}[]
\centering
\scalebox{0.4}{\resizebox{\textwidth}{!}{%
\begin{tabular}{@{}ccccccc@{}}
\cmidrule(l){2-7}
\textbf{} & \multicolumn{3}{c}{\textbf{In-vivo}} & \multicolumn{3}{c}{\textbf{Ex-vivo}} \\ \midrule
\textbf{Model} & P & R & F1 & P & R & F1 \\ \midrule
R. Forest & 0.91 & 0.91 & 0.91 & 0.32 & 0.26 & 0.26 \\
XGboost & 0.96 & 0.96 & 0.96 & 0.48 & 0.24 & 0.36 \\
AlexNet & 0.92 & 0.92 & 0.92 & 0.49 & 0.42 & 0.45 \\
VGG-19 & 0.94 & 0.92 & 0.93 & 0.47 & 0.45 & 0.45 \\
Inception & 0.97 & 0.98 & 0.98 & 0.51 & 0.45 & 0.45 \\ \midrule
ProtoNet & 0.71 & 0.72 & 0.72 & 0.70 & 0.70 & 0.70 \\ \midrule
Ours & 0.88 & 0.88 & 0.88 & \textbf{0.82} & \textbf{0.81} & \textbf{0.81} \\ \bottomrule
\end{tabular}}%
}
\caption{Results comparison with previous work. We show the precision (P), recall (R) and F-score (F1) obtained on the  in-vivo and ex-vivo datasets. }
\label{tab:comparison}
\end{table}


\subsection{Discussion}

To have a better understanding of how our model is behaving, we visualize the features embeddings from the image samples from both kidney stones datasets using UMAP \cite{mcinnes2018umap}. We can observe that in the feature space of the in-vivo dataset (Figure \ref{fig:yonathan_imgs}), the WD class overlaps with BR and WW, something previously reported for traditional features in \cite{embc2021}. In the reduced manifold of the ex-vivo dataset (Figure \ref{fig:vincent-estrade_imgs}), we can see that even some of the classes form distinct clusters, classes such as BR and STR tend to lie very close to each other (a similar trend also occurs in classification made by humans) and that overall, the overlap of the WW class is degrading the performance of the model for ex-vivo images. It may be the case that some classes follow a probability distribution, which would need other metric learning method to be used during the model training to alleviate this problem.

It must be emphasized the urologists make use of both surface and section images to perform the classification of a given kidney stones, using Daudon's morpho-constitutional analysis ether under a microscope or with an endoscope \cite{corrales2021}. The results for the shallow and DL models in Table \ref{tab:comparison} are for models trained in both types of images. Our results using meta-learning do not integrate this knowledge, and thus exploring multi-view image fusion \cite{multi-view} represents an interesting area of research to improve our results.

\section{Conclusions}

In this paper, we conducted a study on generalization capabilities of few-shot learning applied to two kidney stones classification. We were interested in evaluating the generalisation capabilities of meta learning-based FSL methods in a scarce data regime (i.e. medical applications). For this purpose, we made use of two datasets with overlapping the same classes but different data distributions: different acquisition conditions, (in-vivo and ex-vivo) and  endoscopy (type, brand, resolution). Where classical machine learning and deep learning models failed to generalize when trained on one dataset and tested on the other one, our few-shot learning method was able to obtain a high performance over of 80\% in accuracy, recall and F1-score. The difference on accuracy from the evaluation of both datasets is of around 8\%, which is much lower compared with the loss on performance from shallow machine and deep learning models. Although we obtained a performance lower than the previous work on the in-vivo dataset, it is still competitive and demonstrate much better generalization capabilities.

\section*{Acknowledgments}
The authors wish to thank the AI Hub and the CIIOT at ITESM for their support for carrying the experiments reported in this paper in their NVIDIA's DGX computer.

{\small
\bibliographystyle{ieee_fullname}
\bibliography{egbib}
}
\end{document}